\documentclass[10pt, a4paper]{article}

\usepackage[final]{lrec2026} 
\usepackage{booktabs}
\usepackage{multirow}
\usepackage{tabularx}
\usepackage{todonotes}
\usepackage{textcomp} 

\usepackage{tcolorbox}
\tcbset{
  promptbox/.style={
    colback=gray!10,
    colframe=black,
    boxrule=0.5pt,
    arc=4pt,
    left=6pt,
    right=6pt,
    top=6pt,
    bottom=6pt
  }
}

\usepackage{graphicx}

\definecolor{mygreen}{HTML}{E0FF91}
\definecolor{myorange}{HTML}{ffd9b3}

\definecolor{myskyblue}{HTML}{87CEEB}   
\definecolor{mylightcoral}{HTML}{F08080}  

\usepackage{listings}

\lstdefinestyle{jsonplain}{
  backgroundcolor=\color{white},
  basicstyle=\ttfamily\small,
  breaklines=true,
  frame=none,
  showstringspaces=false,
}

\title{Physical Commonsense Reasoning for Lower-Resourced Languages and Dialects: a Study on Basque}

\name{Jaione Bengoetxea, Itziar Gonzalez-Dios, Rodrigo Agerri} 

\address{HiTZ Center - Ixa, University of the Basque Country UPV/EHU \\
         \{jaione.bengoetxea,itziar.gonzalezd,rodrigo.agerri\}@ehu.eus\\}

\abstract{
Physical commonsense reasoning represents a fundamental capability of human intelligence, enabling individuals to understand their environment, predict future events, and navigate physical spaces. Recent years have witnessed growing interest in reasoning tasks within Natural Language Processing (NLP). However, no prior research has examined the performance of Large Language Models (LLMs) on non-question-answering (non-QA) physical commonsense reasoning tasks in low-resource languages such as Basque. Taking the Italian GITA as a starting point, this paper addresses this gap by presenting BasPhyCo, the first non-QA physical commonsense reasoning dataset for Basque, available in both standard and dialectal variants. We evaluate model performance across three hierarchical levels of commonsense understanding: (1) distinguishing between plausible and implausible narratives (accuracy), (2) identifying the conflicting element that renders a narrative implausible (consistency), and (3) determining the specific physical state that creates the implausibility (verifiability). These tasks were assessed using multiple multilingual LLMs as well as models pretrained specifically for Italian and Basque. Results indicate that, in terms of verifiability, LLMs exhibit limited physical commonsense capabilities in low-resource languages such as Basque, especially when processing dialectal variants.
 \\ \newline  }

\begin{document}

\maketitleabstract


\section{Introduction}
\label{sec:intro}

Commonsense reasoning represents the human capacity to understand and manipulate real-world objects and their interactions. This domain has attracted considerable attention in Artificial Intelligence research in recent years \cite{DavisSurvey,sun2025survey}. Physical commonsense reasoning, a specific subdomain, addresses events occurring in the physical world by capturing knowledge about everyday objects, their physical properties, and their potential uses and manipulations \cite{bisk2020piqa,pensa2024multi}. 

As a fundamental component of human intelligence, physical commonsense reasoning enables individuals to reason about their environment, anticipate future events, and navigate their surroundings. Recent research has increasingly examined the reasoning capabilities of LLMs, though such investigations have been conducted predominantly in English \cite{bisk2020piqa, storks2021tiered}. 

This paper focuses on Basque, as well as its Western dialect, and Italian, the source language of the dataset upon which our work is based on \cite{pensa2024multi,pensa2024gita4calamita}. These low-resource languages provide valuable insight into LLM performance on complex physical-world reasoning tasks under data-limited conditions.

We manually translated the Italian dataset GITA into standard Basque and automatically adapted it into the Western dialect. The Western dialect was selected due to its peripheral status and documented linguistic distance from other Basque varieties, as established in dialectological research \citep{michelena1981lengua}. This linguistic divergence is corroborated by several NLP studies: \citet{estarrona2023measuring} identified Biscayan (Western) and Souletin as the most distinct among historical Basque dialects, while \citet{bengoetxea-etal-2025-lost} attributed the negative impact of the Western dialect on Natural Language Inference (NLI) performance to its distance from Standard Basque.

We evaluate model performance across three hierarchical reasoning tasks: (i) distinguishing plausible from implausible narratives (accuracy), (ii) identifying conflicting sentences within implausible narratives (consistency), and (iii) determining the physical states that render narratives implausible (verifiability). Our evaluation uses two multilingual LLMs alongside two Italian-pretrained models and one Basque-pretrained model, thereby examining current LLM knowledge of the physical world and human-object interactions.

To our knowledge, this represents the first investigation combining physical commonsense reasoning with Basque dialectal variation. Data and code are publicly available\footnote{\url{https://github.com/hitz-zentroa/BasPhyCo}}. Our investigation presents the following contributions:





\begin{itemize}

\item The first publicly available non-QA physical commonsense reasoning dataset in Basque, including the first such dataset in a Basque dialect (Western).
\item The first evaluation of LLM performance on non-QA physical commonsense reasoning in a low-resource language such as Basque. Results indicate that, in terms of verifiability, LLMs exhibit
limited physical commonsense capabilities in low-resource languages such as Basque, especially when considering dialectal variants.
\item A comprehensive evaluation of LLMs' knowledge gaps when faced with physical commonsense reasoning for low-resource languages shows that this task is still challenging. Additionally, results with Basque language variation show that models pretrained for the target language seem to have a better ability to handle linguistic variation. 
\item Fine-grained evaluation of physical states indicates that models have a general difficulty in correctly predicting these labels, Location, Edible, and Conscious states being particularly challenging. 
\end{itemize}

\section{Related Work}
\label{sec:related-work}

\paragraph{Physical Commonsense}
Recent research has tried to test physical commonsense knowledge of current LLMs. To this end, researchers have developed various datasets and benchmarks, including textual information \cite{rajani-etal-2019-explain,bisk2020piqa,rajani-etal-2020-esprit,storks2021tiered, aroca-ouellette-etal-2021-prost,wang-etal-2023-newton,pensa2024multi,jeong2025everyday}, images \cite{bakhtin2019phyre,hong2021ptr,liu2022things,meng2024phybench}, and videos \cite{weihs2022benchmarking,yu2022pacs, motamed2025generative}. 

Datasets focusing on textual information have been generally presented as Question-Answering (QA) tasks, such as PIQA \citep{bisk2020piqa}. Some works have attempted to introduce other methodologies, such as TRIP \citep{storks2019commonsense}, which is a physical commonsense reasoning benchmark composed of five-sentence stories. It evaluates models on three tasks: classifying stories as plausible or implausible, detecting the conflicting sentence, and identifying the physical states of objects involved.






The majority of the datasets in physical commonsense reasoning have been curated in English. Some exceptions include GITA \citep{pensa2024multi} for Italian, a non-QA physical commonsense reasoning dataset based on TRIP, and EPiK \cite{jeong2025everyday} for Korean, which follows the PIQA dataset, while culturally adapting it to Korean. 


To our knowledge, the sole existing resource for physical commonsense reasoning in Basque is a professionally translated version of the PIQA dataset \citep{baucells-etal-2025-iberobench}, which provides only the validation subset. Consequently, no Basque-language dataset for physical commonsense reasoning exists beyond the question-answering (QA) format.

\begin{table*}
\begin{tabular}{p{2.2cm}|p{2.2cm}|p{2.2cm}|p{2.2cm}|p{2.2cm}|p{2.2cm}}

\toprule 
Type &  Sentence 1 & Sentence 2 & Sentence 3 &  Sentence 4 &  Sentence 5 \\
\midrule
Plausible &  George filled the glass with water. & George put the glass in the microwave.&  George turned on the microwave. & The water got hot. & George put a tea bag in the water.  \\
 Implausible (order) &    George put the glass in the microwave. & George filled the glass with water. & George turned on the microwave.&  The water got hot. &George put a tea bag in the water.  \\
 Implausible (cloze) &  
George filled the glass with water. & George put the glass in the microwave. & George turned on the microwave.& The water got cold.& George put a tea bag in the water  \\
\bottomrule
\end{tabular}
\caption{Example of a story with its plausible and implausible versions.}
\label{tab:example}
\end{table*}


\paragraph{Dialects and Reasoning}
Regarding the use of dialects in commonsense reasoning, \citet{lin2025assessing} have very recently analyzed LLMs’ dialect robustness and fairness with Standardized English (SE)\footnote{We use the terms the authors use in their papers.} and African American Vernacular English (AAVE). They create the ReDial (Reasoning with Dialect Queries)  dataset, a high-quality, end-to-end human-annotated SE-AAVE parallel dataset for reasoning tasks (algorithm, logic, math, and integrated reasoning) that contains over 1.2K parallel prompts in SE and in AAVE. An evaluation on LLM families (GPT, Claude, Llama, Mistral, Phi) shows lower performance when using dialectal prompts.

\citet{pan2025analyzing} analyze dialectal bias on reasoning tasks through a multiple-choice question answering task, where they compare results in Standard American English (SAE)\footnotemark[\value{footnote}] with results in 5 English dialects, such as Chicano, African American, or Indian English. The dataset was generated by applying grammatical perturbations to the original SAE multiple-choice benchmark using the Multi-VALUE package \citep{ziems-etal-2023-multi}. Results demonstrate that dialectal variation was the main reason for accuracy reductions of up to 20\%.


\paragraph{Variation in Basque}

Modern Basque dialects have been studied and categorized into a comprehensive representation of features by \citet{zuazu2008}. In NLP, early works introduced a morpho-syntactically annotated corpus of Basque historical texts as an aid in the normalization process \citep{estarrona-etal-2020-dealing}. Additionally, a corpus of syntactic variation of northern Basque dialects has been presented \citep{uria2012hizkeren}. More recently, \citet{bengoetxea-etal-2025-lost} presented the first manually created modern Basque dialect dataset for the evaluation of Natural Language Inference (NLI).


Finally, Basque dialects have also been considered in some dialectal benchmark works such as \citet{alam-etal-2024-codet} and \citet{FAISAL2024DIALECTBENCHAN}, which presented benchmarks for Machine Translation (MT) with northern Basque dialects.


\section{Data}
\label{sec:data}



This study examines physical commonsense reasoning in Italian and Basque. We employed GITA \citep{pensa2024multi}, an Italian dataset derived from TRIP \citep{storks2019commonsense}, and manually translated it into Basque. GITA was selected as the foundation dataset due to its manual construction by a professional linguist with explicit attention to semantic coherence. Additionally, whereas TRIP incorporates compound sentences, GITA consists exclusively of simple sentences. This structural simplification reduces linguistic complexity and potential subjectivity, thereby isolating physical reasoning capabilities from confounding syntactic factors during model evaluation.

The following section introduces GITA and the process of its adaptation into both standard and dialectal Basque. 

\subsection{GITA}

GITA \cite{pensa2024multi} is an  Italian physical commonsense evaluation dataset which consists of 356 5-sentence stories, out of which 117 are plausible, and 239 are implausible. The stories focus on concrete actions of the physical world, while mental actions such as ``to think'' or ``to like'' are avoided. 

Two methods were used to create the implausible stories: (i) \textbf{Order}, where the order of two sentences was switched, and (ii) \textbf{Cloze}, where a plausible sentence has been substituted with an implausible one.




Consequently, individual sentences within the narratives are independently plausible but become implausible when placed with specific sentences in implausible narrative sequences. This design ensures that the reasoning task requires the use of the entire context.

In Table \ref{tab:example} we present an example translated into English. The plausible line contains the story with the logical sequence of events. In the implausible (order) example, the order of sentence 1 and 2 has been switched to make a non-logical and implausible story, and in the implausible (cloze) example, the adjective in sentence 3 has been changed from the original \textit{hot} to \textit{cold}, which makes no logical sense as the microwave heats water up.


\subsection{BasPhyCo}

BasPhyCo is the first non-QA physical commonsense reasoning dataset for Basque, available in both standard and dialectal variants. BasPhyCo has been created by manually translating GITA from Italian to Standard Basque.

The translation process included a localization phase in which two linguists adapted cultural elements of GITA to align with Basque contexts. These adaptations included proper names and references to local meteorological agencies, among other culturally-specific elements. The translations adhered closely to standard Basque conventions, specifically excluding lexical items characteristic of Basque dialectal variants.







\subsection{BasPhyCo\textsubscript{west}}



The Standard Basque dataset was automatically converted to Western Basque using a few-shot prompting strategy implemented with the Latxa-3.1-Instruct model \cite{Sainz2025InstructingLL}. Western Basque was selected for two reasons: (1) as a peripheral dialect, it exhibits substantial linguistic distance from Standard Basque, making it a valuable subject for comparative analysis; and (2) preliminary experiments with LLM-based automatic adaptation of GITA revealed a consistent tendency towards Western Basque generation. This methodology leveraged Latxa's perceived tendency to generate Western dialect while accounting for its perceived divergence from standard Basque. The conversion prompt is provided in Appendix \ref{sec:adaptation_prompt}. 





Given that plausible and implausible story pairs contain identical sentences (with the exception of one sentence in cloze implausible narratives), the adaptation process grouped each plausible story with all corresponding implausible variants. The conversion prompt explicitly instructed consistent adaptation of repeated sentences across variants. This methodology ensured uniformity in the adapted narratives.



An example of this adaptation can be found in Table \ref{tab:dialectal-example}, where words like \textit{lorategia} (garden) have been adapted to its Western form \textit{lorategixa}, as well as auxiliary verbs such as \textit{du} have been adapted to \textit{dau}. 

\begin{table}[!t]
    \begin{small}
    \centering
    \begin{tabularx}{\columnwidth}{X|X} \toprule
        Standard & Dialectal \\ 
        \midrule
        Jonek lorategi handi bat \colorbox{myskyblue}{dauka.}
        Elur \colorbox{myskyblue}{lorategian} dago.
        Jonek lorategiko \colorbox{myskyblue}{atea ireki du.}
        Elurrek \colorbox{myskyblue}{alde} egin \colorbox{myskyblue}{du.} 
        \colorbox{myskyblue}{Lorategia} hutsik dago. 

        & 

        Jonek lorategi handi bat \colorbox{mylightcoral}{dauko}.
        Elur \colorbox{mylightcoral}{lorategixen} dago.
        Jonek lorategiko \colorbox{mylightcoral}{atie zabaldu dau}.
        Elurrek \colorbox{mylightcoral}{ospa} egin \colorbox{mylightcoral}{dau.}
        \colorbox{mylightcoral}{Lorategixe} hutsik dago. \\  \bottomrule
    \end{tabularx}
    \caption{Example of a story adapted from Standard Basque to Western Basque.}
    \label{tab:dialectal-example}
    \end{small}
\end{table}


A native professional linguist validated the automatic adaptations to assess overall quality (including minor formatting issues mitigated through prompt engineering) and identify dialectal adaptation errors. The subsequent subsections detail the findings from this initial manual inspection.

However, further evaluation of the quality of this automatic evaluation would be required in the future, in order to assess dialectal adaptation abilities of LLMs.

\subsubsection{Correct Adaptations}

During the manual evaluation step, different types of dialectal linguistic modifications were identified. 

\paragraph{Lexical features} Some lexical changes found to correspond to the Western dialect include \textit{itzali} > \textit{amatatu} (to switch off), \textit{galtzak} > \textit{prakak} (trousers) or \textit{jolas egin} > \textit{olgetan egin} (to play), to name a few. Not only that, but many words have also displayed Western phonology features, such as \textit{ordulari\textbf{XE}} > \textit{ordulari\textbf{A}} (clock) or \textit{salda} > \textit{saldea} (soup). 



\paragraph{Morphosyntactic features} Some common Western morphosyntactic characteristics include the comitative (\textit{norekin}, with what/who) case marker, which in Standard Basque is marked with -KIN, while in the Western dialect this case is represented with the termination -GAZ, as in the following example: \textit{aterkiare\textbf{KIN}} > \textit{aterkixe\textbf{GAZ}} (with the umbrella). 

In terms of auxiliary verb forms, the majority of them have been adapted into the Western dialect, such as \textit{da} > \textit{dau}, \textit{nuen} > \textit{neban}, \textit{ditut} > \textit{dodaz}, to name but a few. 







\subsubsection{Incorrect Adaptations} 

During the manual inspection of the adapted dialectal sentences, we found the following errors. 

\paragraph{Lexical deviations} Some sentences contained made-up words that looked like dialectal words, such as \textit{mugikorra} (phone, standard) > \textit{*mobillora} or \textit{tomate} (tomato, standard) > \textit{*totame}. These lexical adaptations are not part of the Western dialectal vocabulary and could be considered examples of model hallucination. However, they represent a very minimal part of the whole dataset. 

Additionally, some words contained changes that mimic Western dialectal phonology (e.g. \textit{baten} > \textit{*paten}, \textit{sagar} > \textit{*saga}), but are not in fact a part of Western dialectal phonological changes. 


\paragraph{Morphosyntactic deviations} Although sentences generally follow dialectal morphosyntactic patterns, some outputs are not aligned with known dialectal features. For instance, some sentences with missing or additional ergative markers were found: the sentence \textit{*Teknikarixa\textbf{K} ez dau oraindiño etorri}\footnote{Translation: the technician has not arrived yet.}, has an extra ergative marker -K, as intransitive verbs do not need this marker.

Additionally, some sentences contained verb concordance mismatches, such as \textit{*\textbf{indiolarrak} hartu \textbf{dau}}\footnote{Translation: [someone] has taken the turkey.}, where the noun the verb is referring to is plural, but the verb form is singular. Thus, the preferred form would be \textit{\textbf{indiolarrak} hartu \textbf{dauz}}


The observed morphosyntactic divergences are not attested in dialectal corpora, suggesting they stem from the model's generalization errors rather than dialectal norms.





\begin{figure}
    \centering
    \includegraphics[width=\linewidth]{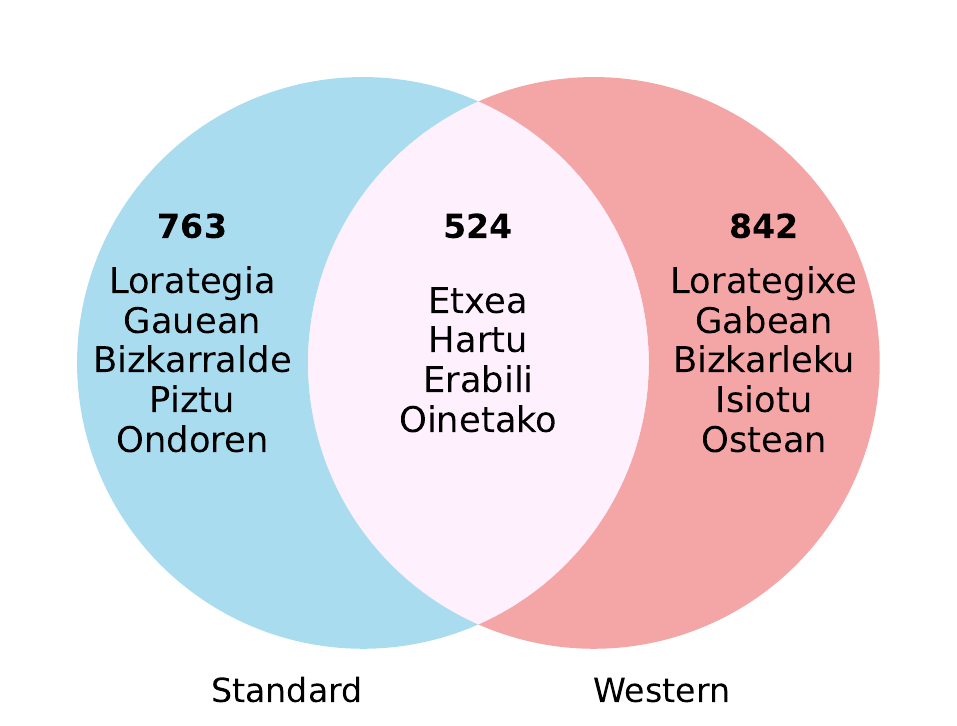}
    \caption{The number of unique words in BasPhyCo (left) and BasPhyCo\textsubscript{west} (right), as well as the overlap of both datasets (middle). Additionally, some examples from each dataset.}
    \label{fig:dataset-analysis}
\end{figure}

In Figure \ref{fig:dataset-analysis}, we illustrate differences and similarities between the Standard and Western Basque datasets, highlighting both their lexical overlaps and divergences. While a portion of the vocabulary is shared between the two varieties, the analysis reveals that there is a substantial part of the lexicon that differs. We additionally present a series of contrastive examples that exemplify the most salient lexical and orthographic variations across the datasets.

\section{Experimental Setup}
\label{sec:experimental-setting}

This section presents the three evaluated tasks and their associated metrics, followed by a description of the selected models and evaluation framework.

\subsection{Task description}
\label{sec:task_description}

Our setup is based on GITA4CALAMITA, a GITA version which was adapted to work with generative LLMs for the CALAMITA shared task \citep{pensa2024gita4calamita}. This approach evaluated three different tasks, which were based on the mirroring of human reasoning, from the shallowest to the deepest. The evaluated tasks are the following:


\begin{itemize}
 \item \textbf{Story classification} determines if the story is plausible or not. Continuing with the example in Table \ref{tab:example}, the plausible story should be classified as plausible and the other two as implausible.
 
\item \textbf{Conflict detection} involves identifying sentence pairs where the story becomes implausible. The conflicting sentences in the example of implausible-order in Table \ref{tab:example} are sentences 1 and 2, since once George puts the glass in the microwave, it is not logical to fill it with water.

\item  \textbf{Physical state classification} recognizes the involved physical states in the conflicting sentences of implausible stories. In the case of the example implausible-cloze in  Table \ref{tab:example}, the involved physical state is the temperature.
\end{itemize}

As in GITA4CALAMITA, we restrict the  physical states to 14: location, conscious, dressed, wet, exist, clean, power, functional, in pieces, open, temperature, solid, occupied, and edible. 

\subsubsection{Data Annotation}

We adopt the annotation from \citet{pensa2024gita4calamita}, which was manually revised by a professional linguist. Some minor annotation errors were detected and corrected, such as occasional mislabeling between \emph{cloze} and \emph{order} story types.

The following is an example from the dataset and its annotation. Some relevant fields include \textit{Type}, which can be \emph{Null} for plausible stories, and \emph{Order} or \emph{Cloze} for implausible ones;\textit{ Confl\_sents} and \textit{Confl\_pairs}, that indicate which sentences make the story implausible.

\begin{small}

\begin{lstlisting}[style=jsonplain]
{
  "0-C0": {
    "story_id": 0,
    "type": "cloze",
    "sentences": [
     "Mikelek hozkailua ireki du.", 
     "Mikelek esnea hartu du.", 
     "Mikelek katilua hartu du.", 
     "Mikelek goilara hartu du.", 
     "Mikelek goilara katiluan sartu du."
    ],
    "length": 5,
    "example_id": "0-C0",
    "plausible": false,
    "breakpoint": 1,
    "confl_sents": [0],
    "confl_pairs": [0, 1]
  }
}
\end{lstlisting}
\end{small}




\begin{table*}
\begin{small}

\centering

\begin{tabular}{llrrr}
\toprule
Test data & Model & Accuracy & Consistency & Verifiability \\   \midrule
\multirow{8}{*}{GITA} &      Llama-3.1-8B-It     & 72.47 &  29.29 & 14.23 \\
                  &         Llama-3.1-70B-It    & \textbf{87.64} & \textbf{65.27} & 36.40  \\ 
                  &         Gemma-2-9B-It       & 71.91 & 35.98 & 16.74  \\
                  &         Gemma-2-27B-It      & 67.42 & 30.96 & 15.06  \\ \cmidrule{2-5}
                  &         Latxa-3.1-8B-It     & 67.42 & 23.85 & 13.39  \\
                  &         Latxa-3.1-70B-It    & 85.96 & 60.25 & \textbf{40.17}  \\ \cmidrule{2-5}
                  &         Minerva-7B-It       & 38.20 & 1.67  & 0.00   \\
                  &         LlaMAntino-3-8B-It  & 58.99 & 18.41 & 7.95   \\ \midrule \midrule
\multirow{8}{*}{BasPhyCo} &       Llama-3.1-8B-It     & 57.30 & 11.30 & 5.02   \\
                  &         Llama-3.1-70B-It    & \textbf{84.83} & 47.70 & 26.78  \\ 
                  &         Gemma-2-9B-It       & 66.01 & 25.10 & 7.53   \\
                  &         Gemma-2-27B-It      & 62.36 & 24.27 & 8.37   \\ \cmidrule{2-5}
                  &         Latxa-3.1-8B-It     & 65.45 & 23.43 & 8.79   \\
                  &         Latxa-3.1-70B-It    & 81.46 & \textbf{48.12} & \textbf{30.54}  \\ \cmidrule{2-5}
                  &         Minerva-7B-It       & 36.52 & 2.09  & 0.42   \\
                  &         LlaMAntino-3-8B-It  & 37.64 & 3.77  & 1.67   \\ \midrule \midrule
\multirow{8}{*}{BasPhyCo\textsubscript{west}} &   Llama-3.1-8B-It     & 51.12 & 9.62  & 4.18   \\
                  &         Llama-3.1-70B-It    & 74.16 & 35.56 & 17.57  \\ 
                  &         Gemma-2-9B-It       & 64.61 & 21.34 & 5.44   \\
                  &         Gemma-2-27B-It      & 57.87 & 18.41 & 6.69   \\ \cmidrule{2-5}
                  &         Latxa-3.1-8B-It     & 63.48 & 17.99 & 9.21   \\
                  &         Latxa-3.1-70B-It    & \textbf{80.34} & \textbf{46.86} & \textbf{28.03}  \\ \cmidrule{2-5}
                  &         Minerva-7B-It       & 38.48 & 2.51  & 0.42   \\
                  &         LlaMAntino-3-8B-It  & 36.24 & 2.93  & 1.67   \\ \bottomrule
                  
\end{tabular}

\caption{Overall results for Story Classification, Conflict Detection and Physical State Classification, measured by accuracy, consistency and verifiability, respectively. GITA: original Italian data; BasPhyCo: manually translated Standard Basque data; BasPhyCo\textsubscript{west}: automatically adapted data into the Western dialect.}
\label{tab:results_overall}

\end{small}
\end{table*}

\subsection{Metrics}

To evaluate model performance, we adopt a tiered evaluation framework \citep{storks2021tiered, pensa2024gita4calamita}. In this setup, each task is evaluated conditionally on the success of the previous one, forming a crescendo of increasingly demanding reasoning requirements. Specifically, only the correctly solved instances from one level are used as input to the next. Accordingly, we adopt three complementary metrics for the three evaluated tasks:


\begin{itemize}

\item \textbf{Accuracy}: Quantifies the proportion of the correctly identified plausible and implausible stories. This metric will be used in the story classification task.

\item \textbf{Consistency}: Measures the proportion of the correctly identified plausible sentences and the conflicting ssentencesin the implausible stories. This measure aims to check the models' consistency when recognizing conflicts. Thus, this metric will be used to evaluate the conflict detection task.

\item  \textbf{Verifiability}: Evaluates the proportion of the correctly identified plausible sentences, the conflicting sentence and the underlying physical states. This shows that the detected conflict can be validated because the underlying implausible change of physical states has been correctly understood. This last metric will be used to evaluate the physical state classification task.
\end{itemize}

\subsection{Evaluation Setup}

We have evaluated our task on generative models, as previous works that evaluated discriminative models \citep{storks2019commonsense, pensa2024multi} were outperformed by generative models \citep{pensa2024gita4calamita}. The evaluation for the three tasks is implemented on EleutherAI's Language Model Evaluation Harness framework v0.4.9 \citep{eval-harness}. This system enables the evaluation of generative LLMs and tasks in a reproducible, automated, and systematic way. The experiments were carried out in a 3-example few-shot setting specified by Harness. All code and prompts are publicly available\footnote{\url{https://github.com/hitz-zentroa/BasPhyCo}}. 





We evaluated all tasks across the three test datasets representing Italian and Standard and Western Basque (Section \ref{sec:data}). The evaluation employed four multilingual models, Llama-3.1 of 8B and 70B parameters \cite{dubey2024llama} and Gemma-2 9B and 27B parameters \cite{team2024gemma}, alongside language-specific models pretrained on Italian (Minerva-7B \cite{orlando-etal-2024-minerva}  and LlaMAntino-3-8B \cite{Polignano2024AdvancedNI}) and Basque, namely, Latxa-3.1-8B and Latxa-3.1-70B \cite{Sainz2025InstructingLL}. All models were instruction-tuned variants.

\begin{table*}
\begin{small}

\centering
\begin{tabular}{ll|rrr|rr|rr}
\toprule
\multirow{2}{*}{Language} & \multirow{2}{*}{Model} & \multicolumn{3}{c|}{Accuracy} & \multicolumn{2}{c|}{Consistency} & \multicolumn{2}{c}{Verifiability} \\ \cmidrule{3-9}
 &   &  Order  &  Cloze  &  Plausible  &  Order  &  Cloze   &  Order  &  Cloze \\ \cmidrule{1-9}
\multirow{8}{*}{GITA} &      Llama-3.1-8B-It     & 64.75 & 81.20 & 71.79 & 14.75 & 44.44 & 4.92  & 23.93 \\
                  &         Llama-3.1-70B-It    & \textbf{88.52} & \textbf{95.73} & \textbf{78.63} & \textbf{55.74} & 74.36 & \textbf{25.41} & 47.01 \\ 
                  &         Gemma-2-9B-It       & 49.18 & 81.20 & 86.32 & 15.57 & 57.26 & 2.46  & 31.62 \\
                  &         Gemma-2-27B-It      & 34.43 & 78.63 & 90.60 & 9.02  & 53.85 & 4.10  & 26.50 \\ 
                  \cmidrule{2-9}
                  &         Latxa-3.1-8B-It     & 55.74 & 68.38 & \textbf{78.63} & 13.11 & 35.04 & 6.56  & 20.51 \\
                  &         Latxa-3.1-70B-It    & 84.43 & 94.87 & \textbf{78.63} & 45.90 & \textbf{75.21} & \textbf{25.41} & \textbf{55.56} \\ \cmidrule{2-9}
                  &         Minerva-7B-It       & 13.11 & 13.68 & 88.89 & 0.00  & 3.42  & 0.00  & 0.00  \\
                  &         LlaMAntino-3-8B-It  & 38.52 & 61.54 & 77.78 & 7.38  & 29.91 & 1.64  & 14.53 \\ \midrule \midrule
\multirow{8}{*}{BasPhyCo} &       Llama-3.1-8B-It     & 46.72 & 50.43 & 75.21 & 8.20  & 14.53 & 2.46  & 7.69  \\
                  &         Llama-3.1-70B-It    & \textbf{79.51} & \textbf{88.03} & \textbf{87.18} & 31.97 & \textbf{64.10} & \textbf{18.03} & 35.90 \\ 
                  &         Gemma-2-9B-It       & 51.64 & 72.65 & 74.36 & 13.11 & 37.61 & 3.28  & 11.97 \\
                  &         Gemma-2-27B-It      & 40.16 & 70.94 & 76.92 & 9.84  & 39.32 & 2.46  & 14.53 \\ 
                  \cmidrule{2-9}
                  &         Latxa-3.1-8B-It     & 48.36 & 67.52 & 81.20 & 10.66 & 36.75 & 4.10  & 13.68 \\
                  &         Latxa-3.1-70B-It    & 76.23 & 87.18 & 81.20 &\textbf{ 36.07} & 60.68 & \textbf{18.03} & \textbf{43.59} \\ \cmidrule{2-9}
                  &         Minerva-7B-It       & 11.48 & 16.24 & 82.91 & 0.82  & 3.42  & 0.00  & 0.85  \\
                  &         LlaMAntino-3-8B-It  & 6.56  & 12.82 & 94.87 & 0.00  & 7.69  & 0.00  & 3.42  \\ \midrule \midrule
\multirow{8}{*}{BasPhyCo\textsubscript{west}} &  Llama-3.1-8B-It  & 34.43 & 44.44 & 75.21 & 4.10  & 15.38 & 1.64  & 6.84  \\
                  &          Llama-3.1-70B-It   & 67.21 & 76.92 & \textbf{78.63} & 20.49 & 51.28 & 6.56  & 29.06 \\ 
                  &         Gemma-2-9B-It       & 66.39 & 68.38 & 58.97 & 13.93 & 29.06 & 3.28  & 7.69  \\
                  &         Gemma-2-27B-It      & 42.62 & 59.83 & 71.79 & 6.56  & 30.77 & 1.64  & 11.97 \\ 
                  \cmidrule{2-9}
                  &          Latxa-3.1-8B-It    & 50.00 & 64.96 & 76.07 & 8.20  & 28.21 & 3.28  & 15.38  \\
                  &         Latxa-3.1-70B-It    & \textbf{78.69} & \textbf{83.76} & \textbf{78.63} & \textbf{32.79} & \textbf{61.54} & \textbf{17.21} & \textbf{39.32}  \\ \cmidrule{2-9}
                  &         Minerva-7B-It       & 22.13 & 17.95 & 76.07 & 1.64  & 3.42  & 0.82  & 0.00   \\
                  &         LlaMAntino-3-8B-It  & 4.92  & 8.55  & 96.58 & 0.00  & 5.98  & 0.00  & 3.42  \\ \bottomrule
                  
\end{tabular}

\caption{Fine-grained examples for all three metrics. GITA: original Italian data; BasPhyCo: manually translated Standard Basque data; BasPhyCo\textsubscript{west}: automatically adapted data into the Western dialect.}
\label{tab:results_detail}

\end{small}
\end{table*}

\section{Results}
\label{sec:results}

We present the results for the three tasks in Table \ref{tab:results_overall}, for Italian (GITA), Standard Basque (BasPhyCo) and Western Basque (BasPhyCo\textsubscript{west}).




\paragraph{Italian} 



The multilingual Llama-3.1-70B-It model achieved the highest performance in accuracy and consistency metrics, while Latxa-3.1-70B-It outperformed other models in terms of verifiability. Conversely, Italian-pretrained models (Minerva-7B-It and LlaMAntino-3-8B-It) yielded the lowest performance across all evaluated tasks, with Minerva-7B-It showing notably inferior results compared to LlaMAntino-3-8B-It.

Notably, Basque-trained Latxa models outperformed Italian-specific models when evaluated on Italian data. Specifically, the smaller Latxa-8B-It model, despite being comparable in size to the Italian models, consistently surpassed LlaMAntino-3-8B-It across all tasks. This performance advantage can be attributed to Latxa's continual pretraining approach \cite{etxaniz-etal-2024-latxa}, which effectively mitigates catastrophic forgetting from its base model, Llama-2.

\paragraph{Standard Basque}
While Llama-3.1-70B-It obtained the highest accuracy score for story classification (84.83 vs 81.46), the Basque pretrained model Latxa-3.1-70B-It had higher scores for the other two more fine-grained metrics, consistency (47.70 vs 48.12) and verifiability (26.78 vs 30.54), respectively. 







\paragraph{Western Basque} Latxa-3.1-70B-It obtained the highest results across all metrics. Llama's performance drop from standard to dialectal data is worth mentioning, as all three metrics undergo important drops (84.83 vs 74.16 for accuracy, 47.70 vs 35.56 for consistency, 26.78 vs 17.57 for verifiability). With Latxa, although there is a performance drop from standard to dialectal, the drop is not nearly as dramatic (81.46  vs 80.34, 48.12 vs 46.86, 30.54 vs 28.03).  These results highlight the importance of pretraining in the target language, as it appears to facilitate more fine-grained linguistic competence and enhance robustness to language variation.








\paragraph{Overall}

LLMs demonstrate notably poor performance in predicting \emph{verifiable} instances for low-resource languages, with performance degrading further when applied to dialectal data.
Regarding task-specific performance, Llama-3.1-70B-It exhibited optimal results in story classification for Italian and Standard Basque, whereas Latxa-3.1-70B-It demonstrated superior consistency and \emph{verifiability}, particularly for Standard and Western Basque. These results indicate that pretraining on target language data yields more substantial improvements in complex reasoning tasks. Additionally, Latxa-3.1-70B-It achieved the highest performance in \emph{verifiability}, which is the most cognitively demanding reasoning task across all evaluated languages.



Finally, the drop from the shallowest to the deepest reasoning task for all models is to be highlighted. Table \ref{tab:results_overall} shows substantial performance degradation, especially in the physical state classification task (verifiability). These findings indicate that, although some models are able to identify implausible stories, providing explanations for their implausibility presents a considerably more challenging task. This will be further discussed in Section \ref{sec:discussion}.

\section{Discussion}
\label{sec:discussion}

In this section, we focus on more fine-grained results, as the three metrics have been specifically computed for the different types of implausible stories (order and cloze). This analysis aims to identify any possible biases that the models could have towards implausible story types.

The results for all three metrics, as well as for the different types of implausible stories, are presented in Table \ref{tab:results_detail}. The main finding indicates that order implausible stories consistently yield lower scores than cloze implausible stories across all metrics, models, and languages. This pattern suggests that the models exhibit stronger reasoning capabilities when confronted with a conflicting sentence within a narrative sequence, compared to cases where implausibility arises solely from the reordering of sentences. These results are consistent with the findings reported by \citet{pensa2024gita4calamita}.








\begin{table}[!t]
\centering
\resizebox{\columnwidth}{!}{%
\begin{tabular}{ll|rrrr}
\toprule
P. state & total & GITA & BasPhyCo  & BasPhyCo\textsubscript{west} & Avrg \\ 
\midrule
Open & 88 & 20.45 & 19.32 &  12.50 & 17.42 \\
Functional & 53 & 26.41 & 15.09 &  16.98 & 19.49 \\
Exist & 47 & 27.66 & 23.40 &  21.28 & 24.11 \\
Power & 36 & 41.67 & 36.11 &  13.63 & 30.47 \\
In pieces & 35 & 42.86  & 25.71 &  31.43 & 33.33 \\
Location & 33 & 15.15 & 6.06  &  6.06 & 9.09\\
Edible & 22 & 4.54 & 0.00 &  4.54 & 3.03 \\
Conscious & 13 & 7.69 & 7.69 &  15.38 & 10.25 \\
Temperature & 12 & 50.00 & 50.00 &  41.67 & 47.22 \\
Wet & 7 & 42.86 & 58.57 &  14.28 & 38.57 \\
Solid & 5 & 80.00 & 60.00 & 40.00 & 60.00\\
Wearing & 3 & 0.00 & 0.00 & 0.00 & 0.00\\
Clean & 1 & 0.00 & 0.00 & 0.00 & 0.00 \\
Occupied & 1 & 100.00 & 100.00  &  100.00 & 100.00\\ 
\bottomrule
\end{tabular}%
}
\caption{Verifiability results per physical state. These results are for Latxa-3.1-70B-It, the model with the highest verifiability results for Italian, Standard and Western Basque.}
\label{tab:physical-state-verif}
\end{table}

Italian and Standard Basque seem to follow similar patterns. Llama-3.1-70B-It obtains the highest results in the majority of the tasks and story types, only being surpassed by Latxa-3.1-70B-It in consistency and verifiability cloze story types. This suggests that, for Italian and Standard Basque, while Llama obtains higher results in shallower reasoning tasks (story classification), Latxa seems to perform slightly better in reasoning tasks involving physical state classification (verifiability).

Regarding the results for Western Basque, Latxa-3.1-70B-It outperforms all other models, including both multilingual models and those pretrained for Italian, following general results in Table \ref{tab:results_overall}. 


Furthermore, the general decrease in performance observed for Llama-3.1-70B-It compared to the standard Basque results highlights the need for multilingual language models that could better handle Basque dialectal variation.

Finally, Latxa-3.1-70B-It consistently obtains high verifiability results for both order and cloze  types, which is the metric that measures how much physical states are predicted correctly. This suggests Latxa's capacity to deal with deeper reasoning tasks such as physical state classification.




\paragraph{Per Physical State Label Verifiability} 

In Table \ref{tab:physical-state-verif}, we report the verifiability results for each physical state label across Italian, as well as Standard and Western Basque. Labels represented by fewer than ten instances are excluded from the following analysis, due to potential sampling bias. Consequently, the subsequent analysis focuses exclusively on those physical states with sufficient representation (i.e., more than ten instances), ensuring more reliable and interpretable results.

Overall, the findings indicate that no particular physical state is consistently easier to predict than the others. In general, performance across categories remains relatively low, highlighting both the intrinsic complexity of this reasoning task and the current limitations of LLMs in capturing nuanced physical state distinctions.

The predictions of Location, Edible, and Conscious states appear to be particularly challenging, as reflected by their comparatively lower verifiability scores. These results suggest that such categories may involve subtleties that LLMs struggle to capture effectively, possibly due to their dependence on implicit world knowledge.

\section{Conclusion}
\label{sec:conclusion}


This paper introduces a novel dataset for evaluating physical commonsense reasoning in Basque and its Western dialect. The dataset was derived from GITA, a manually curated Italian corpus, which underwent manual translation and localization into Standard Basque. Subsequently, the Standard Basque data was automatically adapted to the Western Basque dialect, followed by manual post-editing to ensure accuracy and minimize errors.

We have carried out a suite of experiments to see how multilingual and language-specific LLMs perform on the tasks of physical commonsense reasoning. To our knowledge, this is the first evaluation of non-QA physical commonsense reasoning in low-resourced languages such as Basque and its dialectal varieties. To that end, we have followed a tiered strategy with three tasks of different depth levels: story classification, conflict detection and physical state classification. The results show the LLM's ability to predict verifiable instances is generally low, which highlights the need for further research in the field of physical commonsense reasoning. Further analysis has indicated that identifying implausible instances is more complex when the only change is sentence order. Finally, physical state classification remains a particularly challenging task. 


This work establishes a baseline evaluation framework for commonsense reasoning in low-resource languages and dialectal varieties. Future research directions include extending the dataset to additional languages and dialects.


\section*{Limitations}




The physical commonsense reasoning dataset we present in this work may be culturally localized, reflecting the norms and logic of certain communities, and may need to be adapted to other cultures to be applicable in other contexts.

Additionally, the size of our dataset is currently limited. Expanding the test data and building a training set could alleviate this issue.

Finally, it is important to recognize the inherent bias of Basque LLMs toward Western Basque. Current models show a strong tendency to generate Western Basque features, indicating that their training data and modeling are heavily aligned with this dialect. Expanding this ability to other dialects could enable the analysis of additional variations.


\section*{Acknowledgments}

This work has been supported by the HiTZ center and the Basque Government (Research group funding IT-1805-22).
Jaione Bengoetxea is funded by the Basque Government pre-doctoral grant (PRE\_2024\_1\_0028).

We also acknowledge the following MCIN/AEI/10.13039/501100011033 project: (i) DeepMinor (CNS2023-144375) and European Union NextGenerationEU/PRTR and (ii) DeepThought (PID2024-159202OB-C21) funded by ERDF, EU.

\section*{Bibliographical References}\label{sec:reference}

\bibliographystyle{lrec2026-natbib}
\bibliography{lrec2026-example}

@inproceedings{bisk2020piqa,
  title={{PIQA: Reasoning about Physical Commonsense in Natural Language}},
  author={Bisk, Yonatan and Zellers, Rowan and Gao, Jianfeng and Choi, Yejin and others},
  booktitle={Proceedings of the AAAI conference on artificial intelligence},
  volume={34},
  number={05},
  pages={7432--7439},
  year={2020}
}

@article{storks2019commonsense,
  title={{Commonsense Reasoning for Natural Language Understanding: A Survey of Benchmarks, Resources, and Approaches}},
  author={Storks, Shane and Gao, Qiaozi and Chai, Joyce Y},
  journal={arXiv preprint arXiv:1904.01172},
  pages={1--60},
  year={2019}
}

@inproceedings{storks2021tiered,
  title={{Tiered Reasoning for Intuitive Physics: Toward Verifiable Commonsense Language Understanding}},
  author={Storks, Shane and Gao, Qiaozi and Zhang, Yichi and Chai, Joyce},
  booktitle={Findings of the Association for Computational Linguistics: EMNLP 2021},
  pages={4902--4918},
  year={2021}
}

@inproceedings{pensa2024gita4calamita,
  title={{GITA4CALAMITA-Evaluating the Physical Commonsense Understanding of Italian LLMs in a Multi-layered Approach: A CALAMITA Challenge}},
  author={Pensa, Giulia and Azurmendi, Ekhi and Etxaniz, Julen and Altuna, Bego{\~n}a and Gonzalez-Dios, Itziar},
  booktitle={Proceedings of the 10th Italian Conference on Computational Linguistics (CLiC-it 2024)},
  pages={1153--1160},
  year={2024}
}

@inproceedings{pensa2024multi,
  title={{A Multi-layered Approach to Physical Commonsense Understanding: Creation and Evaluation of an Italian Dataset}},
  author={Pensa, Giulia and Altuna, Bego{\~n}a and Gonzalez-Dios, Itziar},
  booktitle={Proceedings of the 2024 Joint International Conference on Computational Linguistics, Language Resources and Evaluation (LREC-COLING 2024)},
  pages={819--831},
  year={2024}
}

@article{sun2025survey,
  title={{A Survey of Reasoning with Foundation Models: Concepts, Methodologies, and Outlook}},
  author={Sun, Jiankai and Zheng, Chuanyang and Xie, Enze and Liu, Zhengying and Chu, Ruihang and Qiu, Jianing and Xu, Jiaqi and Ding, Mingyu and Li, Hongyang and Geng, Mengzhe and others},
  journal={ACM Computing Surveys},
  volume={57},
  number={11},
  pages={1--43},
  year={2025},
  publisher={ACM New York, NY}
}

@article{DavisSurvey,
author = {Davis, Ernest},
title = {{Benchmarks for Automated Commonsense Reasoning: A Survey}},
year = {2023},
issue_date = {April 2024},
publisher = {Association for Computing Machinery},
address = {New York, NY, USA},
volume = {56},
number = {4},
issn = {0360-0300},
journal = {ACM Comput. Surv.},
month = oct,
articleno = {81},
}

@inproceedings{lin2025assessing,
  title={{Assessing Dialect Fairness and Robustness of Large Language Models in Reasoning Tasks}},
  author={Lin, Fangru and Mao, Shaoguang and La Malfa, Emanuele and Hofmann, Valentin and de Wynter, Adrian and Wang, Xun and Chen, Si-Qing and Wooldridge, Michael J and Pierrehumbert, Janet and Wei, Furu},
  booktitle={Proceedings of the 63rd Annual Meeting of the Association for Computational Linguistics (Volume 1: Long Papers)},
  pages={6317--6342},
  year={2025}
}

@article{pan2025analyzing,
  title={{Analyzing Dialectical Biases in LLMs for Knowledge and Reasoning Benchmarks}},
  author={Pan, Eileen and Choi, Anna Seo Gyeong and ter Hoeve, Maartje and Seto, Skyler and Koenecke, Allison},
  journal={arXiv preprint arXiv:2510.00962},
  year={2025}
}

@article{jeong2025everyday,
  title={{Everyday Physics in Korean Contexts: A Culturally Grounded Physical Reasoning Benchmark}},
  author={Jeong, Jihae and Lee, DaeYeop and Lee, DongGeon and Yu, Hwanjo},
  journal={arXiv preprint arXiv:2509.17807},
year={2025}
}

@inproceedings{rajani-etal-2020-esprit,
    title = "{{ESPRIT}: Explaining Solutions to Physical Reasoning Tasks}",
    author = "Rajani, Nazneen Fatema  and
      Zhang, Rui  and
      Tan, Yi Chern  and
      Zheng, Stephan  and
      Weiss, Jeremy  and
      Vyas, Aadit  and
      Gupta, Abhijit  and
      Xiong, Caiming  and
      Socher, Richard  and
      Radev, Dragomir",
    editor = "Jurafsky, Dan  and
      Chai, Joyce  and
      Schluter, Natalie  and
      Tetreault, Joel",
    booktitle = "Proceedings of the 58th Annual Meeting of the Association for Computational Linguistics",
    year = "2020",
    address = "Online",
    publisher = "Association for Computational Linguistics",
    pages = "7906--7917"
}

@inproceedings{rajani-etal-2019-explain,
    title = "{Explain Yourself! Leveraging Language Models for Commonsense Reasoning}",
    author = "Rajani, Nazneen Fatema  and
      McCann, Bryan  and
      Xiong, Caiming  and
      Socher, Richard",
    editor = "Korhonen, Anna  and
      Traum, David  and
      M{\`a}rquez, Llu{\'i}s",
    booktitle = "Proceedings of the 57th Annual Meeting of the Association for Computational Linguistics",
       year = "2019",
    address = "Florence, Italy",
    publisher = "Association for Computational Linguistics",
         pages = "4932--4942"
}

@inproceedings{wang-etal-2023-newton,
    title = "{{NEWTON}: Are Large Language Models Capable of Physical Reasoning?}",
    author = "Wang, Yi  and
      Duan, Jiafei  and
      Fox, Dieter  and
      Srinivasa, Siddhartha",
    editor = "Bouamor, Houda  and
      Pino, Juan  and
      Bali, Kalika",
    booktitle = "Findings of the Association for Computational Linguistics: EMNLP 2023",
    month = dec,
    year = "2023",
    address = "Singapore",
    publisher = "Association for Computational Linguistics",
    url = "https://aclanthology.org/2023.findings-emnlp.652/",
    doi = "10.18653/v1/2023.findings-emnlp.652",
    pages = "9743--9758",
}

@inproceedings{aroca-ouellette-etal-2021-prost,
    title = "{{PROST}: {P}hysical Reasoning about Objects through Space and Time}",
    author = "Aroca-Ouellette, St{\'e}phane  and
      Paik, Cory  and
      Roncone, Alessandro  and
      Kann, Katharina",
    editor = "Zong, Chengqing  and
      Xia, Fei  and
      Li, Wenjie  and
      Navigli, Roberto",
    booktitle = "Findings of the Association for Computational Linguistics: ACL-IJCNLP 2021",
        year = "2021",
    address = "Online",
    publisher = "Association for Computational Linguistics",
      pages = "4597--4608"
}

@article{hong2021ptr,
  title={{PTR: A Benchmark for Part-based Conceptual, Relational, and Physical Reasoning}},
  author={Hong, Yining and Yi, Li and Tenenbaum, Josh and Torralba, Antonio and Gan, Chuang},
  journal={Advances in Neural Information Processing Systems},
  volume={34},
  pages={17427--17440},
  year={2021}
}

@article{bakhtin2019phyre,
  title={{PHYRE: A New Benchmark for Physical Reasoning}},
  author={Bakhtin, Anton and van der Maaten, Laurens and Johnson, Justin and Gustafson, Laura and Girshick, Ross},
  journal={Advances in Neural Information Processing Systems},
  volume={32},
  year={2019}
}

@inproceedings{liu2022things,
  title={{Things not Written in Text: Exploring Spatial Commonsense from Visual Signals}},
  author={Liu, Xiao and Yin, Da and Feng, Yansong and Zhao, Dongyan},
  booktitle={Proceedings of the 60th Annual Meeting of the Association for Computational Linguistics (Volume 1: Long Papers)},
  pages={2365--2376},
  year={2022}
}

@article{meng2024phybench,
  title={{PhyBench: A Physical Commonsense Benchmark for Evaluating Text-to-Image Models}},
  author={Meng, Fanqing and Shao, Wenqi and Luo, Lixin and Wang, Yahong and Chen, Yiran and Lu, Quanfeng and Yang, Yue and Yang, Tianshuo and Zhang, Kaipeng and Qiao, Yu and others},
  journal={CoRR},
  year={2024}
}

@article{motamed2025generative,
  title={{Do Generative Video Models Understand Physical Principles?}},
  author={Motamed, Saman and Culp, Laura and Swersky, Kevin and Jaini, Priyank and Geirhos, Robert},
  journal={arXiv preprint arXiv:2501.09038},
  year={2025}
}

@inproceedings{yu2022pacs,
  title={{PACS: A Dataset for Physical Audiovisual CommonSense Reasoning}},
  author={Yu, Samuel and Wu, Peter and Liang, Paul Pu and Salakhutdinov, Ruslan and Morency, Louis-Philippe},
  booktitle={European Conference on Computer Vision},
  pages={292--309},
  year={2022},
  organization={Springer}
}

@article{weihs2022benchmarking,
  title={{Benchmarking Progress to Infant-level Physical Reasoning in AI}},
  author={Weihs, Luca and Yuile, Amanda and Baillargeon, Ren{\'e}e and Fisher, Cynthia and Marcus, Gary and Mottaghi, Roozbeh and Kembhavi, Aniruddha},
  journal={Transactions on Machine Learning Research},
  year={2022}
}

@inproceedings{ziems-etal-2023-multi,
    title = "{Multi-{VALUE}: A Framework for Cross-Dialectal {E}nglish {NLP}}",
    author = "Ziems, Caleb  and
      Held, William  and
      Yang, Jingfeng  and
      Dhamala, Jwala  and
      Gupta, Rahul  and
      Yang, Diyi",
    editor = "Rogers, Anna  and
      Boyd-Graber, Jordan  and
      Okazaki, Naoaki",
    booktitle = "Proceedings of the 61st Annual Meeting of the Association for Computational Linguistics (Volume 1: Long Papers)",
    month = jul,
    year = "2023",
    address = "Toronto, Canada",
    publisher = "Association for Computational Linguistics",
    url = "https://aclanthology.org/2023.acl-long.44/",
    doi = "10.18653/v1/2023.acl-long.44",
    pages = "744--768"
}

@article{michelena1981lengua,
  title={Lengua com{\'u}n y dialectos vascos},
  author={Mitxelena, Luis},
  journal={Anuario del Seminario de Filolog{\'\i}a Vasca" Julio de Urquijo"},
  volume={15},
  pages={289--313},
  year={1981}
}

@article{estarrona2023measuring,
  title={{Measuring Language Distance for Historical Texts in Basque}},
  author={Estarrona, Ainara and Etxeberria, Izaskun and Padilla-Moyano, Manuel and Soraluze, Ander},
  journal={Procesamiento del Lenguaje Natural},
  volume={70},
  pages={53--61},
  year={2023}
}

@inproceedings{bengoetxea-etal-2025-lost,
    title = "{Lost in Variation? Evaluating {NLI} Performance in {B}asque and {S}panish Geographical Variants}",
    author = "Bengoetxea, Jaione  and
      Gonzalez-Dios, Itziar  and
      Agerri, Rodrigo",
    editor = "Boleda, Gemma  and
      Roth, Michael",
    booktitle = "Proceedings of the 29th Conference on Computational Natural Language Learning",
    month = jul,
    year = "2025",
    address = "Vienna, Austria",
    publisher = "Association for Computational Linguistics",
    url = "https://aclanthology.org/2025.conll-1.30/",
    doi = "10.18653/v1/2025.conll-1.30",
    pages = "452--468",
    ISBN = "979-8-89176-271-8"
}

@book{zuazu2008,
    title = {Euskalkiak. Euskararen dialektoak},
    author = {Zuazu, Koldo},
    isbn = {978-84-90272-38-1},
    year = {2008},
    publisher = {Elkar}
}

@inproceedings{estarrona-etal-2020-dealing,
    title = "{Dealing with Dialectal Variation in the Construction of the {B}asque Historical Corpus}",
    author = "Estarrona, Ainara  and
      Etxeberria, Izaskun  and
      Etxepare, Ricardo  and
      Padilla-Moyano, Manuel  and
      Soraluze, Ander",
    editor = {Zampieri, Marcos  and
      Nakov, Preslav  and
      Ljube{\v{s}}i{\'c}, Nikola  and
      Tiedemann, J{\"o}rg  and
      Scherrer, Yves},
    booktitle = "Proceedings of the 7th Workshop on NLP for Similar Languages, Varieties and Dialects",
    month = dec,
    year = "2020",
    address = "Barcelona, Spain (Online)",
    publisher = "International Committee on Computational Linguistics (ICCL)",
    url = "https://aclanthology.org/2020.vardial-1.8/",
    pages = "79--89"
}

@article{uria2012hizkeren,
  title={Hizkeren arteko aldakortasun sintaktikoa aztertzeko metodologiaren nondik norakoak: BASYQUE aplikazioa},
  author={Uria, Larraitz and Etxepare, Ricardo},
  journal={Lapurdum. Euskal ikerketen aldizkaria| Revue d'{\'e}tudes basques| Revista de estudios vascos| Basque studies review},
  number={16},
  pages={117--135},
  year={2012},
  publisher={IKER UMR 5478}
}

@inproceedings{alam-etal-2024-codet,
    title = "{{CODET}: A Benchmark for Contrastive Dialectal Evaluation of Machine Translation}",
    author = "Alam, Md Mahfuz Ibn  and
      Ahmadi, Sina  and
      Anastasopoulos, Antonios",
    editor = "Graham, Yvette  and
      Purver, Matthew",
    booktitle = "Findings of the Association for Computational Linguistics: EACL 2024",
    month = mar,
    year = "2024",
    address = "St. Julian{'}s, Malta",
    publisher = "Association for Computational Linguistics",
    url = "https://aclanthology.org/2024.findings-eacl.125/",
    pages = "1790--1859"
}

@article{FAISAL2024DIALECTBENCHAN,
  title={{DIALECTBENCH: A NLP Benchmark for Dialects, Varieties, and Closely-Related Languages}},
  author={Fahin Faisal and Orevaoghene Ahia and Aarohi Srivastava and Kabir Ahuja and David Chiang and Yulia Tsvetkov and Antonios Anastasopoulos},
  journal={ArXiv},
  year={2024},
  volume={abs/2403.11009},
  url={https://api.semanticscholar.org/CorpusID:268513057}
}

@inproceedings{baucells-etal-2025-iberobench,
    title = "{{I}bero{B}ench: A Benchmark for {LLM} Evaluation in {I}berian Languages}",
    author = "Baucells, Irene  and
      Aula-Blasco, Javier  and
      de-Dios-Flores, Iria  and
      Paniagua Su{\'a}rez, Silvia  and
      Perez, Naiara  and
      Salles, Anna  and
      Sotelo Docio, Susana  and
      Falc{\~a}o, J{\'u}lia  and
      Saiz, Jose Javier  and
      Sepulveda Torres, Robiert  and
      Barnes, Jeremy  and
      Gamallo, Pablo  and
      Gonzalez-Agirre, Aitor  and
      Rigau, German  and
      Villegas, Marta",
    editor = "Rambow, Owen  and
      Wanner, Leo  and
      Apidianaki, Marianna  and
      Al-Khalifa, Hend  and
      Eugenio, Barbara Di  and
      Schockaert, Steven",
    booktitle = "Proceedings of the 31st International Conference on Computational Linguistics",
    month = jan,
    year = "2025",
    address = "Abu Dhabi, UAE",
    publisher = "Association for Computational Linguistics",
    url = "https://aclanthology.org/2025.coling-main.699/",
    pages = "10491--10519"
}

@misc{eval-harness,
  author       = {Gao, Leo and Tow, Jonathan and Abbasi, Baber and Biderman, Stella and Black, Sid and DiPofi, Anthony and Foster, Charles and Golding, Laurence and Hsu, Jeffrey and Le Noac'h, Alain and Li, Haonan and McDonell, Kyle and Muennighoff, Niklas and Ociepa, Chris and Phang, Jason and Reynolds, Laria and Schoelkopf, Hailey and Skowron, Aviya and Sutawika, Lintang and Tang, Eric and Thite, Anish and Wang, Ben and Wang, Kevin and Zou, Andy},
  title        = {{The Language Model Evaluation Harness}},
  month        = 07,
  year         = 2024,
  publisher    = {Zenodo},
  version      = {v0.4.3},
  doi          = {10.5281/zenodo.12608602},
  url          = {https://zenodo.org/records/12608602}
}

@inproceedings{etxaniz-etal-2024-latxa,
    title = "Latxa: An Open Language Model and Evaluation Suite for {B}asque",
    author = "Etxaniz, Julen  and
      Sainz, Oscar  and
      Perez, Naiara  and
      Aldabe, Itziar  and
      Rigau, German  and
      Agirre, Eneko  and
      Ormazabal, Aitor  and
      Artetxe, Mikel  and
      Soroa, Aitor",
    editor = "Ku, Lun-Wei  and
      Martins, Andre  and
      Srikumar, Vivek",
    booktitle = "Proceedings of the 62nd Annual Meeting of the Association for Computational Linguistics (Volume 1: Long Papers)",
    year = "2024",
    pages = "14952--14972",
}

@article{Sainz2025InstructingLL,
  title={Instructing Large Language Models for Low-Resource Languages: A Systematic Study for Basque},
  author={Oscar Sainz and Naiara P{\'e}rez and Julen Etxaniz and Joseba Fernandez de Landa and Itziar Aldabe and Iker Garc\'ia-Ferrero and Aimar Zabala and Ekhi Azurmendi and Germ{\'a}n Rigau and Eneko Agirre and Mikel Artetxe and Aitor Soroa},
  journal={ArXiv},
  year={2025},
  volume={abs/2506.07597}
}

@article{team2024gemma,
  title={Gemma 2: Improving open language models at a practical size},
  author={Team, Gemma and Riviere, Morgane and Pathak, Shreya and Sessa, Pier Giuseppe and Hardin, Cassidy and Bhupatiraju, Surya and Hussenot, L{\'e}onard and Mesnard, Thomas and Shahriari, Bobak and Ram{\'e}, Alexandre and others},
  journal={arXiv preprint arXiv:2408.00118},
  year={2024}
}

@article{dubey2024llama,
  title={The llama 3 herd of models},
  author={Dubey, Abhimanyu and Jauhri, Abhinav and Pandey, Abhinav and Kadian, Abhishek and Al-Dahle, Ahmad and Letman, Aiesha and Mathur, Akhil and Schelten, Alan and Yang, Amy and Fan, Angela and others},
  journal={arXiv preprint arXiv:2407.21783},
  year={2024}
}

@inproceedings{orlando-etal-2024-minerva,
    title = "Minerva {LLM}s: The First Family of Large Language Models Trained from Scratch on {I}talian Data",
    author = "Orlando, Riccardo  and
      Moroni, Luca  and
      Huguet Cabot, Pere-Llu{\'i}s  and
      Conia, Simone  and
      Barba, Edoardo  and
      Orlandini, Sergio  and
      Fiameni, Giuseppe  and
      Navigli, Roberto",
    editor = "Dell'Orletta, Felice  and
      Lenci, Alessandro  and
      Montemagni, Simonetta  and
      Sprugnoli, Rachele",
    booktitle = "Proceedings of the Tenth Italian Conference on Computational Linguistics (CLiC-it 2024)",
    year = "2024",
    publisher = "CEUR Workshop Proceedings",
    pages = "707--719",
}

@article{Polignano2024AdvancedNI,
  title={Advanced Natural-based interaction for the ITAlian language: LLaMAntino-3-ANITA},
  author={Marco Polignano and Pierpaolo Basile and Giovanni Semeraro},
  journal={ArXiv},
  year={2024},
  volume={abs/2405.07101}
}





\appendix

\section{Automatic Adaptation Prompt}
\label{sec:adaptation_prompt}

Figure \ref{fig:prompt} provides the prompt that was used to obtain the standard-to-dialectal adaptations of the Basque dataset.

\newpage
\clearpage

\begin{small}
    
\begin{figure*}[!b]
\centering
\begin{tcolorbox}[promptbox]

I will give you three versions of a story. Each version has five sentences. Some sentences are identical across versions. You need to adapt this text so that it includes Bizkaian dialectal features. You can use non-standard orthography. Try to make it as similar as possible to oral language.

Task:

\begin{enumerate}
    \item First, list all unique sentences across all three stories.
    \item Adapt each unique sentence exactly once into the Bizkaian dialect.
    \item Then reconstruct the three stories with the translations, making sure that any identical source sentence always has the identical translation.
    \item If there are more than three stories, repeat the same process for all of them.
\end{enumerate}

Format:

This is an example of an standard (INPUT) instance and an example of the dialectal (OUTPUT) adaptation that you need to do:

Standard:

STORY1: ['Mikel lanera joan da', 'Mikelek ordenagailua piztu du', 'Mikelek mezuak irakurri ditu', 'Mikelek mezuak erantzun ditu', 'Mikel etxera joan da']

STORY2: ['Mikel lanera joan da', 'Mikelek mezuak erantzun ditu', 'Mikelek mezuak irakurri ditu', 'Mikelek ordenagailua piztu du', 'Mikel etxera joan da']

STORY3: ['Mikel lanera joan da', 'Mikelek ordenagailua itzali du', 'Mikelek mezuak irakurri ditu', 'Mikelek mezuak erantzun ditu', 'Mikel etxera joan da']

Dialectal:

STORY1: ['Mikel lanera jun de', 'Mikelek ordenagaillua piztu dau', 'Mikelek mesuek irakurri dauz', 'Mikelek mesuek erantzun dauz', 'Mikel etxera jun de']

STORY2: ['Mikel lanera jun de', 'Mikelek mesuek erantzun ditu', 'Mikelek mesuek irakurri dauz', 'Mikelek ordenagaillua piztu dau', 'Mikel etxera jun de']

STORY3: ['Mikel lanera jun de', 'Mikelek ordenagaillua amatatu dau', 'Mikelek mesuek irakurri dauz', 'Mikelek mesuek erantzun dauz', 'Mikel etxera jun de']

Output only the reconstructed stories in the exact same format as the input. Do not output explanations, steps, or commentary.
\end{tcolorbox}
\caption{Dialectal adaptation prompt.}
\label{fig:prompt}
\end{figure*}

\end{small}

\end{document}